\documentclass{article}

\usepackage[numbers]{natbib}

\usepackage{booktabs}

\usepackage[dblblindworkshop,final]{neurips_2025}

\workshoptitle{MATH-AI}

\usepackage[utf8]{inputenc} % allow utf-8 input
\usepackage[T1]{fontenc}    % use 8-bit T1 fonts
\usepackage{hyperref}       % hyperlinks
\usepackage{url}            % simple URL typesetting
\usepackage{booktabs}       % professional-quality tables
\usepackage{amsfonts}       % blackboard math symbols
\usepackage{nicefrac}       % compact symbols for 1/2, etc.
\usepackage{microtype}      % microtypography
\usepackage{xcolor}         % colors

\title{Systematic Diagnosis of Brittle Reasoning in Large Language Models}

\author{
  V. S. Raghu Parupudi\thanks{Use footnote for providing further information
    about author (webpage, alternative address)---\emph{not} for acknowledging
    funding agencies.} \\
  University of California, San Diego\\
  La Jolla, CA 92092 \\
  \texttt{pvsrrkishore@gmail.com} \\
  \texttt{s1parupudi@ucsd.edu.com} \\
}

\begin{document}

\maketitle

\begin{abstract}
  A central question in artificial intelligence is the extent to which machine learning models comprehend mathematics. To address this, we propose a novel framework for measuring mathematical reasoning that moves beyond standard benchmarks to diagnose specific failure points. Our method first generates structured, step-by-step reasoning from gpt-3.5-turbo on the GSM8K dataset. We then use a more capable analyst model, gpt-4o-mini, to categorize errors and, crucially, perform an unsupervised clustering of every reasoning sentence to identify emergent "reasoning modes." This analysis reveals a cognitive profile with a stark, nonhuman-like brittleness: while the model achieves near-perfect accuracy on procedural modes like sequential calculation, its performance on modes requiring combinatorial reasoning with restrictions plummets. By identifying and quantifying the reliability of these distinct reasoning skills, our work provides a more granular method to evaluate mathematical comprehension and offers a precise roadmap for developing new capabilities and more reliable future applications.
\end{abstract}

\section{Introduction}

The ability of Large Language Models (LLMs) to perform mathematical reasoning has advanced dramatically, driven by two key developments. First, the creation of high-quality benchmarks, such as the GSM8K dataset, provided a clear target to measure multi-step reasoning skills \cite{cobbe2021gsm8k}. Second, the discovery of Chain-of-Thought (CoT) prompting demonstrated that eliciting a series of intermediate steps could unlock reasoning capabilities in these models, significantly boosting performance beyond even supervised methods of the time \cite{wei2022chain}.

These advancements have shifted the research frontier from merely achieving correct final answers to ensuring the faithfulness and reliability of the reasoning process itself. The most promising efforts in this area have focused on process supervision, which provides feedback for each intermediate reasoning step during training. As Lightman et al. (2023) demonstrated, this approach is significantly more effective than supervising only the final outcome, leading to more robust models \cite{lightman2023verify}. However, their work also highlights a critical, unresolved issue: even state-of-the-art, process-supervised models still regularly produce logical mistakes. While the field has developed powerful methods for training models to reason, we lack a systematic, scalable framework for diagnosing the failures that persist. When a model fails, is it due to a miscalculation, a flawed logical inference, or a deeper misinterpretation of the problem?

This paper introduces a novel framework for the automated, post-hoc diagnosis of these reasoning failures. Our work complements training-focused approaches by providing a method to measure and understand the specific cognitive patterns that lead to errors. Our pipeline first generates structured reasoning traces from a generator model (gpt-3.5-turbo) and then uses a more capable analyst model (gpt-4o-mini) to identify and categorize the first point of failure in incorrect solutions.

Our primary contribution is the discovery, through unsupervised clustering, of distinct "reasoning modes" and the quantification of their reliability. This analysis reveals a cognitive profile of LLM reasoning that is both powerful and profoundly brittle.

\section{Related Work}
 The literature on the mathematical reasoning capabilities of LLMs can be broadly organized into two key themes: techniques for generating and supervising reasoning paths, and the use of LLMs themselves as tools for analysis and evaluation.

\subsection{Generating and Supervising Reasoning Paths}

The dominant paradigm for eliciting mathematical reasoning is the Chain of Thought (CoT) method, which prompts models to produce a series of intermediate steps leading to a final answer \cite{wei2022chain}. This approach has been shown to be highly effective on benchmarks like GSM8K \cite{cobbe2021gsm8k}. Recognizing the limitations of a single, linear reasoning path, subsequent research has explored more complex reasoning structures. The "Tree of Thoughts" (ToT) framework, for instance, generalizes CoT by allowing a model to explore multiple, divergent reasoning paths and self-evaluate choices, significantly enhancing performance on problems that require strategic planning or search \cite{yao2023tree}.

Concurrent with efforts to improve reasoning generation, a major focus has been on improving model reliability through fine-grained supervision. The work on process supervision, notably by Lightman et al. (2023), demonstrated that providing feedback for each intermediate reasoning step is significantly more effective for training reliable models than supervising only the final outcome \cite{lightman2023verify}. This has established that the quality of intermediate steps is a critical factor in achieving robust performance.

\subsection{The LLM-as-Evaluator Paradigm}

The scaling of LLM capabilities has led to their use as automated evaluators of other models' outputs. The validity of this "LLM-as-a-Judge" approach was rigorously examined by Zheng et al. (2023), who found that strong models like GPT-4 can achieve over 80\% agreement with human preferences on open-ended tasks, making them a scalable and reliable tool for evaluation \cite{zheng2023judging}. This paradigm has also been applied in iterative refinement loops. The Self-Refine framework, for example, uses a single LLM to generate an initial output, provide feedback on its own work, and then use that feedback to improve its output, demonstrating performance gains across a variety of tasks \cite{madaan2023selfrefine}. These works provide a strong foundation for using LLMs as a core component in an analytical workflow.

\section{Experimental Setup and Methodology}

\subsection{Experimental Setup}
Our experiments were conducted on a random sample of 1,000 problems from the training split of the GSM8K dataset \cite{cobbe2021gsm8k}, selected with a fixed random seed for reproducibility. We utilized three models from the OpenAI API, all with temperature set to 0.0 to ensure deterministic outputs. A gpt-3.5-turbo-1106 model served as the "generator" to produce initial reasoning traces. OpenAI's text-embedding-3-large was used to generate sentence embeddings. A more capable gpt-4o-mini model served as the "analyst" for subsequent diagnostic and auto-labeling tasks.

\subsection{Methodology}
Our methodology begins with structured reasoning elicitation, where we prompted the generator model to return solutions in a JSON format containing a step-by-step reasoning trace and a final answer. This forces the model's reasoning into a discrete, machine-readable format. Next, in the automated diagnosis stage, the analyst model programmatically examined each failed trace to identify and categorize the first point of failure.

The core of our method is the analysis of reasoning modes via clustering. We performed an unsupervised clustering on every sentence from all generated traces. This process involved several steps. First, each sentence was converted into a high-dimensional vector using the text-embedding-3-large model. These embeddings \cite{reimers2019sentencebert} were then L2-normalized. We then applied the HDBSCAN \cite{campello2013hdbscan} clustering algorithm directly to these high-dimensional normalized embeddings to group them by semantic similarity. The resulting semantic clusters were then automatically labeled by prompting our gpt-4o-mini analyst to generate a concise, descriptive summary for a sample of sentences from each group. This process allowed us to identify emergent "modes" of reasoning (e.g., 'calculating total costs').

To measure the reliability of these modes, we labeled each entire reasoning trace based on its final, verifiable outcome; if the answer was wrong, every sentence in the trace was considered part of a "failed reasoning process." This trace-level labeling is justified because any single error invalidates the entire reasoning, effectively we are punishing the model for taking a confusing route. This method allows for a clean, binary outcome of the trace to serve as a robust statistical signal for our analysis.  Finally, we calculated a "correctness rate" for each cluster, defined as the percentage of its sentences that belonged to a successfully completed reasoning trace. This transforms the sentence clusters into a quantifiable map of the model's skills.

\subsection{Contribution to the Field}

First, this work provides a novel tool for measuring mathematical reasoning that moves beyond task-level accuracy to a granular, diagnostic map of a model's cognitive profile. Second, it offers a new comparative lens on AI vs. human cognition by revealing the non-human-like 'brittleness' where mastery and total failure coexist. Finally, by pinpointing the exact reasoning modes that are brittle, our framework provides a clear, data-driven agenda for future research to build more robust and reliable models

\section{Results and Discussion}

We applied our diagnostic pipeline to 1,000 reasoning traces generated by gpt-3.5-turbo on the GSM8K dataset. Out of the 1,000 problems, the model produced the correct final answer for 849, resulting in an overall accuracy of 84.9\%. Our analysis focuses on the 151 incorrect responses to diagnose the underlying patterns of failure and reliability.

\subsection{High-Level Failure Categorization}
First, our analyst model categorized the first point of failure for each of the 151 incorrect traces. This provides a high-level overview of the error types, with the distribution shown in Table~\ref{tab:failure-summary}. The most frequent cause of failure was Reasoning Error, which occurred in nearly half of all failed traces. This suggests that flaws in the logical plan or strategy are more common than simple arithmetic mistakes or misinterpreting the initial problem statement. However, these broad categories do not fully capture the granular nature of the model's weaknesses, motivating our deeper clustering analysis.

\begin{table}[h!]

\caption{Distribution of failure types at the first erroneous step}
\label{tab:failure-summary}
\centering
\begin{tabular}{lrr}
\toprule
\textbf{Error Category} & \textbf{Count} & \textbf{Percentage} \\
\midrule
Reasoning Error & 75 & 49.7\% \\
Calculation Error & 50 & 33.1\% \\
Misinterpretation Error & 17 & 11.3\% \\
Uncategorized by Analyst & 5 & 3.3\% \\
Factual Invention & 4 & 2.6\% \\
\midrule
\textbf{Total Failures Analyzed} & \textbf{151} & \textbf{100.0\%} \\
\bottomrule
\end{tabular}

\end{table}

\subsection{Identifying Robust and Brittle Reasoning Modes}
Our primary finding comes from the unsupervised clustering of all reasoning sentences. This analysis moves beyond error categorization to reveal a stark contrast between highly reliable reasoning modes and those that are exceptionally brittle. To formally validate our findings, we confirmed that the correctness rate for each selected cluster is statistically significant. We performed a Fisher's Exact Test, comparing each cluster's performance against a baseline sentence-level correctness rate derived from the overall 84.9\% problem-level accuracy. For all highlighted clusters in Table~\ref{tab:cluster-results}, the tests yielded p-values of less than 0.05, confirming that the observed performance is not due to random chance. There are other clusters that have a statistically significant difference in performance, but are omitted for brevity.

The results, presented in Table~\ref{tab:cluster-results}, showcase this performance gap. The model demonstrates near-perfect, statistically significant reliability in well-defined, procedural tasks. For example, clusters related to Calculating total cost of items (Cluster 172) and Sequential Calculation Steps (Cluster 47) both achieved 100\% correctness, indicating a mastery of basic arithmetic and procedural execution.

In stark contrast, performance collapses when reasoning requires handling combinatorial constraints, algebraic abstraction, or complex multi-step logic. Most notably, Cluster 11, which corresponds to Calculating topping combinations with restrictions, exhibits a 0.0\% correctness rate. This, along with other low-performing clusters like Substituting and simplifying equations (Cluster 93, 27.3\% correct), represents a clear, systematic, and statistically significant failure mode.

Apart from insights into where gpt-3.5-turbo fails from a reasoning perspective, this work helps in understanding the difference between human and machine cognition. The cognitive profile revealed here—exhibiting both absolute mastery and total, systematic failure on closely related mathematical concepts—is profoundly non-human-like.  A human student might struggle with a concept, but they would not typically display this extreme binary of 100\% success versus 0\% failure. This discovery of "brittle" reasoning modes provides a more granular method for measuring the boundaries of an LLM's mathematical comprehension and offers a clear, data-driven roadmap for the targeted interventions needed to build more robust and reliable AI reasoners.

\begin{table}[h!]

\caption{Correctness of selected reasoning modes identified via clustering}
\label{tab:cluster-results}
\centering
\begin{tabular}{lrrp{6cm}}
\toprule
\textbf{Cluster ID} & \textbf{Correctness} & \textbf{Sentence Count} & \textbf{Auto-Label (Reasoning Mode)} \\
\midrule
\multicolumn{4}{l}{\textit{Examples of Robust Reasoning Modes}} \\
\addlinespace
172 & 100.0\%* & 26 & Calculating total cost of items \\
47  & 100.0\%* & 22 & Sequential Calculation Steps \\
171 & 95.1\%* & 61 & Calculating total costs or profits \\
\addlinespace
\midrule
\multicolumn{4}{l}{\textit{Examples of Brittle Reasoning Modes}} \\
\addlinespace
80  & 46.2\%* & 13 & Calculating current ages and differences \\
14  & 46.2\%* & 13 & Calculating net effects and time \\
93  & 27.3\%* & 11 & Substituting and simplifying equations \\
60  & 27.3\%* & 11 & Calculate and round time or quantity \\
\textbf{11} & \textbf{0.0\%*} & \textbf{11} & \textbf{Calculating combinations with restrictions} \\
\bottomrule
\end{tabular}
\newline
\small{*All correctness rates are statistically significant (p < 0.05) via Fisher's Exact Test against a baseline of 84.9\%.}

\end{table}

\section{Limitations and Future Work}

For this initial investigation, our analysis is centered on a single generator model (gpt-3.5-turbo) and the widely-used GSM8K dataset. While this provides a generalizable case study,  we acknowledge its limitations that provide clear directions for future work. Due to lack of resources, the work is based on a single generator model (gpt-3.5-turbo) and a single dataset (GSM8K). Furthermore, our diagnosis relies on an LLM analyst, and the accuracy of its judgments warrants further investigation.

To build on this work, our plan is threefold. First, we will conduct a cross-model analysis, applying our pipeline to a suite of state-of-the-art models (e.g., GPT-4, Claude 3, Gemini 1.5) to create a comparative "brittleness profile" for each. Second, we will expand our analysis to more complex reasoning domains. Finally, and most critically, we plan to close the loop from diagnosis to solution. By using the sentences from identified brittle clusters (such as Cluster 11) as a targeted dataset for fine-tuning, we will investigate whether this data-efficient intervention can surgically repair these specific reasoning deficiencies, a key step toward building more robust and reliable models.

\bibliographystyle{plain}
\bibliography{neurips_2025}

\end{document}